\documentclass[11pt]{article}

\usepackage[margin=1in]{geometry}
\usepackage{amsmath,amssymb}
\usepackage{graphicx}
\usepackage{booktabs}
\usepackage{array}
\usepackage{hyperref}
\usepackage{xcolor}
\usepackage{microtype}
\usepackage{parskip}
\usepackage{caption}
\usepackage{subcaption}
\usepackage{enumitem}
\usepackage{natbib}
\usepackage{float}

\hypersetup{
  colorlinks=true,
  linkcolor=blue!70!black,
  citecolor=blue!70!black,
  urlcolor=blue!70!black,
}

\title{\textbf{AutoStan: Autonomous Bayesian Model Improvement via Predictive Feedback}}

\author{
Oliver D\"urr\\
\small Thurgau Institute for Digital Transformation (TIDIT), Switzerland
}

\begin{document}

\maketitle
  
\begin{abstract}
We present AutoStan, a framework in which a command-line interface (CLI) coding agent autonomously builds and iteratively improves Bayesian models written in Stan. The agent operates in a loop, writing a Stan model file, executing MCMC sampling, then deciding whether to keep or revert each change based on two complementary feedback signals: the negative log predictive density (NLPD) on held-out data and the sampler's own diagnostics (divergences, R-hat, effective sample size).
We evaluate AutoStan on five datasets with diverse modeling structures.
On a synthetic regression dataset with outliers, the agent progresses from naive linear regression to a model with Student-$t$ robustness, nonlinear heteroscedastic structure, and an explicit contamination mixture, matching or outperforming TabPFN, a state-of-the-art black-box method, while remaining fully interpretable.
Across four additional experiments, the same mechanism discovers hierarchical partial pooling, varying-slope models with correlated random effects, and a Poisson attack/defense model for soccer.
No search algorithm, critic module, or domain-specific instructions are needed.
This is, to our knowledge, the first demonstration that a CLI coding agent can autonomously write and iteratively improve Stan code for diverse Bayesian modeling problems.
\end{abstract}

\section{Introduction}

Bayesian modeling is, in principle, beautifully simple: you state what you believe about the data, condition on observations, and let probability theory do the rest.
In practice, this simplicity was historically locked behind intractable integrals---until probabilistic programming languages like BUGS~\citep{lunn2009}, JAGS~\citep{jags2003}, and Stan~\citep{stan2017} made arbitrary generative models accessible via MCMC.
Yet even with Stan, the practitioner must still closely monitor and "babysit" the inference: diagnose divergences, check R-hat and effective sample sizes, reparameterize funnels, tune priors, and iterate, a workflow that textbooks like \citet{gelman2006} and \citet{mcelreath2020} have made accessible but not less laborious.
The question we tackle in this paper is whether modern CLI coding agents can automate parts of this tedious cycle.

Karpathy's autoresearch~\citep{karpathy2026} demonstrated that a coding agent with a single scalar reward (validation bits-per-byte) can autonomously improve a neural network training script overnight.
We translate this idea to Bayesian modeling.
Supervised Bayesian inference is a natural fit for autonomous optimization because it admits a principled, model-agnostic evaluation metric like the negative log predictive density (NLPD) or any other strictly proper scoring rule~\citep{gneiting2007}.
NLPD rewards calibrated predictive distributions and cannot be gamed by over- or under-confident predictions, making it an ideal hands-off reward signal.
The agent edits a Stan model file, runs MCMC, observes NLPD on held-out data, and decides whether to keep or revert.
No additional orchestration beyond the CLI agent is needed.

We demonstrate AutoStan on five datasets spanning four statistical paradigms: regression with outliers, hierarchical partial pooling, varying-slope models, and sports modeling.
Each reveals qualitatively different agent behaviors: from immediately writing the correct model (hierarchical) to discovering Student-$t$ robustness for outlier handling and ultimately an explicit contamination mixture (regression).
To our knowledge, this is the first demonstration that a CLI coding agent can autonomously write and improve Stan code.

\section{Method}

\subsection{CLI Coding Agents}
A CLI coding agent is a terminal-based AI assistant that can read and edit files, execute shell commands, and reason about their output in a read--edit--execute loop. In our case, the agent edits Stan model files and executes MCMC via cmdstanpy~\citep{cmdstanpy2024}.
Examples of CLI coding agents include Claude Code~\citep{anthropic2025}, Gemini CLI~\citep{geminicli2025}, Codex CLI~\citep{codexcli2025}, and opencode~\citep{opencode2025}, the last of which is open-source. AutoStan relies on this capability but is not tied to any specific agent: any CLI coding agent that can edit files and execute shell commands can implement the optimization loop.
All experiments reported here use Claude Code (v2.1.86) with Sonnet~4.6 (Model ID claude-sonnet-4-6) and a context window of 200{,}000 tokens~\citep{anthropic2025}. The framework, however, is agent-agnostic and should work with any modern CLI coding agent.

\subsection{Workflow}

The agent is launched with a single prompt:
\begin{quote}
\ttfamily Read program.md for your instructions. Your dataset is datasets/<name>.
\end{quote}

\noindent Two files define the task:

\paragraph{\texttt{program.md}} contains domain-agnostic Bayesian workflow instructions (${\sim}$56 lines; full text in Appendix~\ref{app:prompt}): read the data, inspect the training set, edit \texttt{model.stan}, run evaluation, interpret NLPD and diagnostics, decide whether to keep or revert, repeat.
The instructions mention general strategies (non-centered parameterization, prior tuning, alternative likelihoods) but contain no dataset-specific guidance.

\paragraph{\texttt{dataset.md}} is a short dataset description: column names, data format, the Stan data interface, and the dataset-specific evaluation command (\texttt{python evaluate.py}), which the agent calls to compile the model, run MCMC, and receive the resulting NLPD.
For synthetic datasets, variable names are anonymized; the true generative process is never disclosed.
Examples are given in Appendix~\ref{app:datasets} and at \url{https://github.com/tidit-ch/autostan}.

\subsection{Agent Autonomy}

The agent is free to explore the data before writing code.
In practice, it often computes summary statistics, checks for outliers, examines group structure, or inspects value ranges on its own initiative, behavior that emerges from the agent's statistical reasoning rather than from any script.
The agent then writes a \texttt{model.stan} file with complete freedom over priors, likelihood, parameterization, and model structure.
The only contract: the model must output a \texttt{log\_lik} vector in the \texttt{generated quantities} block which is needed for the evaluation script to compute the NLPD.

\subsection{Evaluation and Reward}

Each dataset has an \texttt{evaluate.py} script (executable but not readable by the agent) that compiles the model via cmdstanpy, runs MCMC (4~chains, 1\,000~post-warmup draws for most datasets; 30\,000 for the large 1D regression, to reduce Monte Carlo noise), computes NLPD from the \texttt{log\_lik} vector (provided in the stan file).
Test data is protected via filesystem permissions and never visible to the agent.
The primary reward is NLPD on held-out data:
\begin{equation}
  \mathrm{NLPD} = -\frac{1}{N_{\mathrm{test}}} \sum_{n=1}^{N_{\mathrm{test}}}
  \log \!\left( \frac{1}{S} \sum_{s=1}^{S} p\!\left(y_n^{\mathrm{test}} \mid \theta^{(s)}\right) \right)\!.
\end{equation}
As a strictly proper scoring rule~\citep{gneiting2007}, the NLPD is uniquely minimized by the true predictive distribution. In contrast to metrics such as accuracy or RMSE, it penalizes both errors in the predictive mean and miscalibrated uncertainty. In principle, any other strictly proper scoring rule (e.g., CRPS) could be used.

The agent halts after 3~consecutive non-improving iterations or 20~total.

\section{Case Study: Regression with Outliers}
\label{sec:case_study}
We examine this case study in detail, as it clearly illustrates the core principles of our approach. Additional examples are provided in Section~\ref{sec:generality}.
\subsection{Setup}
\label{sec:setup}

The agent receives a \texttt{dataset.md} file (see Appendix~\ref{app:datasets}) containing the column names (\texttt{predictor}, \texttt{response}), the Stan data interface, and the evaluation command, plus the one-line dataset description: ``Observations of a continuous predictor and a continuous response.''

The true generative process, which is unknown to the agent, involves three independent sources of difficulty:
\begin{align}
  f(x)      &= 2\sin(1.2x) + 0.3x, \label{eq:dgp_mean}\\[2pt]
  \sigma(x) &= 0.3 + 0.8\exp\!\Bigl[-\tfrac{1}{2}\Bigl(\tfrac{x-3}{1.5}\Bigr)^{2}\Bigr], \label{eq:dgp_sigma}
\end{align}
and ${\sim}6\%$ of \emph{training} observations shifted $\pm10$--$15$ units as extreme outliers.
The test set is drawn from the clean generative process (no outliers), so the oracle NLPD is simply the negative log-density under the true $f(x)$ and $\sigma(x)$; no mixture component is needed.
This oracle represents a lower bound that cannot be reached: beyond the usual parameter uncertainty from finite data, the oracle knows that the test set contains no outliers, an information the agent does not have.
We evaluate on two datasets: large ($n{=}500$ train, 200 test) and small ($n{=}68$ train, 30 test).

\subsection{Model Evolution on the Large Dataset}

\begin{figure}[t]
\centering
\includegraphics[width=\linewidth]{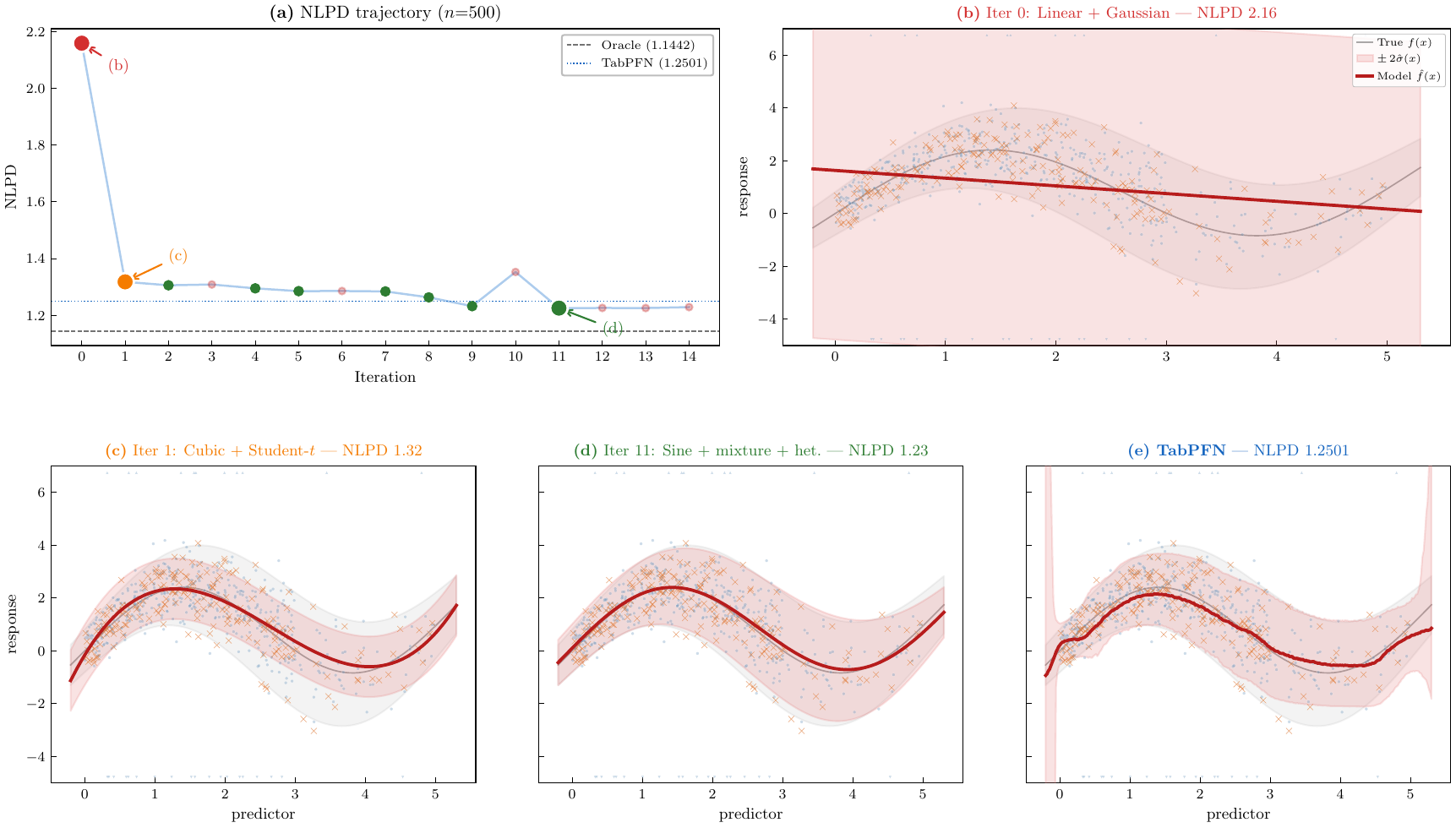}
\caption{%
AutoStan on the large 1D regression dataset ($n{=}500$ train, 200~test;
DGP defined in Section~\ref{sec:setup}).
In panels (b)--(d), the fitted mean $\hat f(x)$ is plotted with $\pm2\hat\sigma(x)$ shaded;
the faint grey overlay marks the oracle ground truth $f(x)\pm2\sigma(x)$.
Training points outside the plot range $[-5,7]$ are shown as edge-pinned triangles
($\blacktriangle$/$\blacktriangledown$) at the panel boundary.
\textbf{(a)}~NLPD trajectory over 15~iterations; colored markers identify the three model-fit panels.
The dashed oracle line ($\mathrm{NLPD}=1.14$) is a lower bound that cannot be reached
(see Section~\ref{sec:setup}).
\textbf{(b)}~Baseline (iter~0): Gaussian linear model; predictive bands are dominated
by the ${\approx}30$ extreme training outliers ($\mathrm{NLPD}=2.16$).
\textbf{(c)}~Iter~1: cubic polynomial mean $+$ Student-$t$ likelihood;
one step eliminates most band inflation ($\mathrm{NLPD}=1.32$).
\textbf{(d)}~Iter~11 (best): sinusoidal mean with learned frequency~$\omega$,
heteroscedastic log-linear variance, and a contamination-mixture likelihood;
tight, locally calibrated bands closely follow the oracle noise envelope ($\mathrm{NLPD}=1.23$).
\textbf{(e)}~TabPFN (90\% predictive interval, $\mathrm{NLPD}=1.25$):
mean tracking is accurate but intervals are uniformly too wide---the model
has absorbed the training outliers, inflating uncertainty across the full input range.%
}
\label{fig:main}
\end{figure}

Figure~\ref{fig:main} summarizes the large-dataset run in one view.
In 15~iterations the agent progresses from a flat, outlier-dominated baseline to a posterior predictive that closely tracks the true DGP (compare panels (b)--(d)), and ends up outperforming TabPFN (panel (e)).

The trajectory reveals four key discoveries (see also Table~\ref{tab:large}):

\paragraph{Discovery 1: Outlier Robustness ($\Delta = 0.84$).}
The baseline linear regression with Gaussian errors scores NLPD~=~2.16.
At iteration~1, the agent adds a cubic polynomial mean and switches to Student-$t$, a single change that drops NLPD by 0.84, the largest jump of the run.

\paragraph{Discovery 2: Nonlinear Mean Structure ($\Delta \approx 0.02$).}
The agent replaces the polynomial with a sinusoidal basis $a\sin(\omega x) + b\cos(\omega x) + cx$ (iter~2).
It subsequently frees $\omega$ as a learnable parameter (iter~7), recovering a value near the true frequency~$1.2$.

\paragraph{Discovery 3: Heteroscedastic Variance ($\Delta \approx 0.02$).}
Adding $\log\sigma(x)$ as a quadratic then cubic polynomial captures the locally elevated variance near $x\approx 3$ (iters~5, 11).

\paragraph{Discovery 4: Robust Mixture Model ($\Delta = 0.05$).}
At iteration~9 the agent replaces Student-$t$ with an explicit two-component Gaussian mixture:
\begin{equation}
  y \sim (1-\pi)\,\mathcal{N}(\mu(x),\, \sigma(x)) + \pi\,\mathcal{N}(\mu(x),\, \sigma_{\mathrm{out}}).
\end{equation}
The posterior estimate $\hat{\pi} \approx 6\%$ is close to the true contamination fraction. However, introducing two exchangeable mixture components leads to a label-switching pathology: the posterior is invariant under permutation of the component labels, effectively swapping the roles of $\sigma(x)$ and $\sigma_{\mathrm{out}}$. As a result, MCMC chains alternate between modes in which either component explains the outliers, leading to poor mixing (R-hat~=~1.52 at iteration~10).
The agent resolves this issue by fixing $\sigma_{\mathrm{out}} = 10$ based on data inspection, thereby restoring stable inference.

\begin{table}[t]
\centering
\setlength{\tabcolsep}{4pt}
\begin{tabular}{clll}
\toprule
\textbf{Iter} & \textbf{NLPD} & \textbf{Model} & \textbf{$\Delta$} \\
\midrule
0 & 2.1589 & Linear mean, Gaussian likelihood & --- \\
1 & 1.3181 & Cubic polynomial $+$ Student-$t$ & $-$0.84 \\
2 & 1.3060 & Sine basis $+$ Student-$t$ & $-$0.01 \\
5 & 1.2854 & $+$ Quadratic $\log\sigma(x)$ & $-$0.02 \\
7 & 1.2844 & $+$ Learnable frequency $\omega$ & $-$0.001 \\
9 & 1.2325 & Gaussian mixture (fixed $\sigma_{\mathrm{out}}{=}10$) & $-$0.05 \\
\textbf{11} & \textbf{1.2256} & \textbf{$+$ Cubic $\log\sigma(x)$} & $\mathbf{-}$\textbf{0.007} \\
\bottomrule
\end{tabular}
\caption{NLPD trajectory, 1D regression large ($n{=}500$). Oracle NLPD~=~1.1442. Full trajectory including rejected iterations in Table~\ref{tab:large_full}.}
\label{tab:large}
\end{table}

\subsection{Small Dataset ($n{=}68$)}

With only $n=68$ training observations, the same three discoveries occur in compressed form (9~iterations; see Appendix~\ref{app:small_traj}):
(1)~outlier robustness via Student-$t$ ($\Delta{=}0.75$),
(2)~nonlinear mean with heteroscedastic variance ($\Delta{=}0.35$), and
(3)~contamination mixture ($\Delta{=}0.03$), reaching NLPD~1.12.
The sinusoidal frequency and cubic heteroscedastic profile, which require more data to identify, are absent here and cannot be identified due to small sample size.

\subsection{Comparison with TabPFN}

We compare against TabPFN~\citep{tabpfn2024}, a transformer pre-trained on synthetic tabular data that claims to approximate Bayesian inference without fitting a model at test time (details in Appendix~\ref{app:tabpfn}).
Other natural baselines exist (gradient-boosted trees~\citep{chen2016} or deep ensembles), but TabPFN is particularly useful out of the box as it produces full predictive distributions rather than point estimates and is currently a state-of-the-art method for regression tasks.
On the small dataset, TabPFN (NLPD~1.12) and AutoStan (NLPD~1.12) are effectively tied; on the large dataset, AutoStan slightly outperforms TabPFN (1.23 vs.\ 1.25).
The gap reflects structural advantages: the sinusoidal mean captures the true functional form, the heteroscedastic variance gives locally calibrated uncertainty, and the mixture correctly separates outliers.
The main advantage, however, is interpretability: AutoStan exposes every assumption (mean, variance profile, outlier fraction $\hat{\pi}$) with posterior uncertainty; TabPFN provides a predictive distribution but no explanation.

\section{Additional Datasets}
\label{sec:generality}
The 1D regression serves as a single illustrative example. To assess how broadly the mechanism generalizes, within the limited scope of this exploratory study, we apply AutoStan to four additional datasets spanning hierarchical, varying-slope, and football (soccer) modeling problems (Table~\ref{tab:summary}; complete trajectories in Appendix~\ref{app:trajectories}).
A large-scale benchmark is beyond the scope of this paper; our goal is instead to demonstrate the range of modeling structures the agent can discover.

\paragraph{Hierarchical: small ($n_j{=}8$, 20 groups).}
The generative structure follows the spirit of the classic Eight Schools model~\citep{rubin1981} (group means drawn from a shared prior, observations drawn from those means), but uses a synthetic dataset with anonymized variable names to avoid LLM contamination.
The agent selected a hierarchical partial-pooling model as its \emph{baseline} (NLPD~=~1.50, oracle: 1.49), reaching near-optimal performance already at iteration~0.
Three attempted improvements (non-centered parameterization (NCP), Student-$t$, group-specific $\sigma$) all increased NLPD.
The finding is the baseline choice itself: given anonymized group-structured data, the agent immediately wrote the correct model.

\paragraph{Hierarchical: large ($n_j{=}40$, 20 groups).}
Same generative process, more data.
The agent achieved four consecutive improvements: NCP, group-specific variances, Student-$t$ likelihood, and tighter data-informed priors, reaching NLPD~1.4014 (oracle: 1.4039).
It also ran a principled ablation, reverting to Normal likelihood to confirm Student-$t$ was beneficial.
However, the best model slightly outperforms the oracle, a result that is not statistically meaningful and likely reflects mild overfitting to the test set through iterative optimization (see Limitations).

\paragraph{Varying slopes ($J{=}15$ groups).}
Generated from correlated random intercepts and slopes, with anonymized variable names.
The largest gain ($\Delta{=}0.51$) came from discovering varying slopes at iter~1.
Most strikingly, the agent discovered a piecewise linear structure \emph{not present in the true generative model} (iter~10), achieving NLPD~1.2738 vs.\ oracle~1.2627.
While the gap is small, the piecewise structure is a modeling artifact rewarded by the test set and is considered another instance of mild test-set adaptation through iterative optimization.
The agent correctly recognized when additional complexity became counterproductive: 3-segment piecewise (15 divergences) and learned knot positions (282 divergences, R-hat~=~1.63) were both rejected.

\paragraph{Bundesliga 2024/25 (real data).}
Real match results, 18 teams, temporal train/test split.
This is the only dataset with domain-labeled column names (\texttt{home\_team\_id}, \texttt{home\_goals}, etc.).
The agent immediately applied a canonical Poisson model with separate attack and defense parameters, then iteratively added hierarchical priors ($\Delta{=}0.020$), NCP, and team-specific home advantage ($\Delta{=}0.003$).
It rejected the negative binomial likelihood, the Dixon--Coles low-scoring correction~\citep{dixon1997}, and a Bradley--Terry-style quality parameter~\citep{bradley1952}, none of which improved NLPD.

An earlier experiment with fully anonymized column names (\texttt{unit\_a}, \texttt{count\_a}) revealed the agent's pattern recognition: from 18 units, small integer counts (0--7), and a paired-matchup structure alone, it reasoned ``\emph{This looks like a sports-style matchup dataset}'' and wrote a Poisson attack/defense model as its baseline, without any domain labels.

\begin{table}[t]
\centering
\setlength{\tabcolsep}{3pt}
\small
\begin{tabular}{lcccccc}
\toprule
 & \textbf{Hier.\ S} & \textbf{Hier.\ L} & \textbf{Slopes} & \textbf{1D S} & \textbf{1D L} & \textbf{Bundes.} \\
\midrule
Oracle & 1.494 & 1.404 & 1.263 & 0.944$^\dagger$ & 1.144$^\dagger$ & --- \\
Baseline & 1.500 & 1.404 & 1.818 & 2.248 & 2.159 & 1.566 \\
Best & \textbf{1.500} & \textbf{1.401} & \textbf{1.274} & \textbf{1.124} & \textbf{1.226} & \textbf{1.543} \\
Gap & 0.006 & $-$0.003 & 0.011 & 0.180$^\dagger$ & 0.081$^\dagger$ & --- \\
Iters & 0 & 4 & 10 & 5 & 11 & 9 \\
\bottomrule
\end{tabular}
\caption{Summary across all experiments. Gap~=~Best~$-$~Oracle (lower is better; negative means the model outperforms the oracle, likely due to test-set overfitting). $^\dagger$Oracle is a strict lower bound that cannot be reached: the oracle knows the test set contains no outliers, whereas the agent must account for the ${\approx}6\%$ training contamination. The agent exhibits qualitatively different behaviors depending on dataset structure and size.}
\label{tab:summary}
\end{table}

\section{Related Work}

A growing body of work explores LLM-driven statistical modeling.
\citet{li2024} structure their system around the classic iterative cycle of proposing a model, fitting it, critiquing the fit, and revising. Implemented in PyMC, an LLM plays both roles: modeler (proposing probabilistic programs) and domain expert (critiquing them).
ModelSMC~\citep{wahl2026} takes a more formal route, recasting model discovery as probabilistic inference with candidate models as particles in a sequential Monte Carlo sampler; theoretically elegant but substantially more complex.
MATHMO~\citep{liu2026} takes a broader scope, treating mathematical modeling in general (not just probabilistic models) as a sequential decision-making problem.
\citet{domke2025} sidesteps iterative improvement entirely: an LLM generates many candidate probabilistic programs from an informal description, and predictions are computed as a weighted average, i.e., Bayesian model averaging with an LLM prior over model space.
At the specification end, LLM-BI~\citep{huang2025} demonstrates that LLMs can elicit priors and likelihoods from natural language, though so far only for linear regression.

All of these approaches introduce dedicated infrastructure: custom critic modules, SMC samplers, model-space priors, or multi-objective search algorithms.
AutoStan is radical in what it \emph{does not} need.
The entire system consists of a generic CLI coding agent, a 56-line instruction file, a short dataset description, and NLPD as the sole reward: no custom framework, no search algorithm, no critic.
Our direct inspiration is Karpathy's autoresearch~\citep{karpathy2026}, which showed that a coding agent with a single scalar reward (validation bits-per-byte) can autonomously improve a neural network training script overnight.
AutoStan translates this to Bayesian modeling, where NLPD (a strictly proper scoring rule) replaces bits-per-byte and Stan model files replace Python training scripts. An extension is that AutoStan augments the scalar reward with the rich diagnostic output of the MCMC sampler (divergences, R-hat, and effective sample sizes), giving the agent a second feedback channel that signals not just whether a model improved but how it is broken and what to fix next.

The contrast with black-box methods is instructive.
TabPFN~\citep{tabpfn2024}, a transformer pre-trained on synthetic datasets, produces competitive predictive distributions, but no model: there are no parameters to interpret, no priors to inspect, no assumptions to challenge.
AutoStan's output is fundamentally different: a complete generative model in Stan with explicit likelihood, priors, and structure that a statistician can read, critique, and build upon.
Moreover, because the agent runs in an interactive CLI, the practitioner can pause the loop, ask the agent to explain its modeling choices in plain language, request specific changes, or resume autonomous improvement; even a user who does not read Stan fluently can ask the agent to translate the model into a verbal description of its assumptions, making the Bayesian workflow accessible well beyond the traditional expert community.

\section{Discussion and Conclusion}

Four findings emerge across all experiments.
\textbf{(1)~NLPD and MCMC diagnostics drive structural discovery}: NLPD signals whether a change improves predictive performance, while the sampler's output (divergences, elevated R-hat, poor mixing) signals \emph{how} a model is broken and what to fix.
Both channels are essential: the agent discovered robust likelihoods, nonlinear heteroscedastic structure, contamination mixtures, hierarchical pooling, varying slopes, and sport-specific models by reading both, and the final output is not just a score but a complete, readable Stan model.
\textbf{(2)~The agent adapts to data structure and scale}: on small hierarchical data it immediately writes the correct model; with more data it finds additional structure (group-specific variances, learned frequencies).
On the 1D regression, 68~observations suffice for outlier robustness; with 500, the agent also recovers the true sinusoidal frequency and cubic heteroscedastic profile.
\textbf{(3)~The agent adapts to domain}: anonymized variable names lead to generic baselines; domain-labeled columns (Bundesliga) lead to domain-appropriate models (Poisson attack/defense).
Even with fully anonymized column names, the agent's world knowledge helps it recognize structural patterns (e.g., paired integer counts as a sports matchup).
\textbf{(4)~The agent knows when to stop}: divergences, elevated R-hat, and increased NLPD serve as reliable stopping signals, and the agent correctly rejects overcomplex models across all datasets.

\textbf{Limitations.}
Iterative NLPD optimization on a fixed test set risks mild overfitting: on the hierarchical-large dataset, the best model slightly outperforms the oracle (NLPD 1.401 vs.\ 1.404), and on varying slopes, the agent discovers piecewise structure not in the true DGP.
These results are not statistically meaningful improvements over the oracle but artifacts of optimizing against a finite test set over multiple iterations.
A mitigation is cross-validation or periodic test-set rotation, though this increases computational cost. When no held-out test set is available, PSIS-LOO \citep{vehtari2017} offers a principled within-sample surrogate: it is computed directly from the \texttt{log\_lik} vector.

Because AutoStan produces inspectable Stan code, a human reviewer can always verify whether a given improvement reflects genuine statistical insight or test-set adaptation, an advantage absent from black-box methods.
A similar issue can arise when the agent uses exploratory data analysis to inform priors. E.g., centering an outlier-scale prior on magnitudes observed in the training data. This technically conflates prior and likelihood, but is acceptable when prediction is the primary goal, as the held-out test set remains an honest judge of predictive quality. For users interested in genuine inference, the agent's output is not a black box but readable Stan code: the data-informed priors can simply be replaced with the analyst's true prior beliefs before a final sampling run.

The agent’s training data presents both advantages and limitations.
On the one hand, it may have encountered similar textbook problems, making it difficult to disentangle genuine statistical reasoning from memorization; we mitigate this by using anonymized synthetic datasets with unpublished generative processes.
On the other hand, domain knowledge is a strength: when the Bundesliga columns carry semantic labels, the agent readily selects an appropriate Poisson attack/defense model, and even with fully anonymized columns it can recover structural patterns from its training.

A CLI coding agent, guided only by NLPD, autonomously discovers sophisticated probabilistic models across diverse datasets (regression, hierarchical, varying-slope, and sports) while producing fully interpretable Stan code.
No search algorithm, critic, or domain-specific tuning is required.
NLPD and MCMC diagnostics are all you need.

Code, datasets, and Stan model snapshots are available at \url{https://github.com/tidit-ch/autostan}. A Claude Code skill for practical use on new data is available at \url{https://github.com/tidit-ch/autostan-skill}.

\pagebreak
\bibliographystyle{plainnat}
\bibliography{references}

@article{vehtari2017,
  author    = {Vehtari, Aki and Gelman, Andrew and Gabry, Jonah},
  title     = {Practical Bayesian model evaluation using leave-one-out cross-validation and {WAIC}},
  journal   = {Statistics and Computing},
  volume    = {27},
  number    = {5},
  pages     = {1413--1432},
  year      = {2017},
  doi       = {10.1007/s11222-016-9696-4},
}

@misc{karpathy2026,
  author       = {Karpathy, Andrej},
  title        = {autoresearch},
  year         = {2026},
  howpublished = {GitHub},
  url          = {https://github.com/karpathy/autoresearch}
}

@article{rubin1981,
  author  = {Rubin, Donald B.},
  title   = {Estimation in parallel randomized experiments},
  journal = {Journal of Educational Statistics},
  year    = {1981},
  volume  = {6},
  number  = {4},
  pages   = {377--401}
}

@inproceedings{li2024,
  author    = {Li, Michael Y. and Fox, Emily B. and Goodman, Noah D.},
  title     = {Automated Statistical Model Discovery with Language Models},
  booktitle = {Proceedings of the 41st International Conference on Machine Learning},
  series    = {Proceedings of Machine Learning Research},
  year      = {2024},
  eprint    = {2402.17879},
  archivePrefix = {arXiv},
  url       = {https://arxiv.org/abs/2402.17879}
}

@misc{wahl2026,
  author        = {Wahl, Stefan and Schenk, Raphaela and Farnoud, Ali and
                   Macke, Jakob H. and Gedon, Daniel},
  title         = {A Probabilistic Framework for {LLM}-Based Model Discovery},
  year          = {2026},
  eprint        = {2602.18266},
  archivePrefix = {arXiv},
  url           = {https://arxiv.org/abs/2602.18266}
}

@inproceedings{liu2026,
  author    = {Liu, Tennison and van der Schaar, Mihaela},
  title     = {{MATHMO}: Automated Mathematical Modeling Through Adaptive Search},
  booktitle = {Proceedings of the 14th International Conference on Learning Representations},
  year      = {2026},
  url       = {https://openreview.net/forum?id=t2fZ2GOwAT}
}

@inproceedings{domke2025,
  author    = {Domke, Justin},
  title     = {Large Language {Bayes}},
  booktitle = {Advances in Neural Information Processing Systems},
  year      = {2025},
  eprint    = {2504.14025},
  archivePrefix = {arXiv},
  url       = {https://arxiv.org/abs/2504.14025}
}

@misc{huang2025,
  author        = {Huang, Yongchao},
  title         = {{LLM-BI}: Towards Fully Automated {B}ayesian Inference with
                   Large Language Models},
  year          = {2025},
  eprint        = {2508.08300},
  archivePrefix = {arXiv},
  url           = {https://arxiv.org/abs/2508.08300}
}

@article{tabpfn2024,
  author    = {Hollmann, Noah and M{\"u}ller, Samuel and Purucker, Lennart and
               Krishnakumar, Arjun and K{\"o}rfer, Max and Hoo, Shi Bin and
               Schirrmeister, Robin Tibor and Hutter, Frank},
  title     = {Accurate predictions on small data with a tabular foundation model},
  journal   = {Nature},
  year      = {2025},
  month     = jan,
  volume    = {637},
  pages     = {319--326},
  doi       = {10.1038/s41586-024-08328-6}
}

@misc{stan2017,
  author       = {{Stan Development Team}},
  title        = {Stan Modeling Language Users Guide and Reference Manual,
                  Version 2.17.0},
  year         = {2017},
  url          = {https://mc-stan.org}
}

@article{gneiting2007,
  author  = {Gneiting, Tilmann and Raftery, Adrian E.},
  title   = {Strictly proper scoring rules, prediction, and estimation},
  journal = {Journal of the American Statistical Association},
  year    = {2007},
  volume  = {102},
  number  = {477},
  pages   = {359--378}
}

@article{lunn2009,
  author  = {Lunn, David and Spiegelhalter, David and Thomas, Andrew and Best, Nicky},
  title   = {The {BUGS} project: Evolution, critique and future directions},
  journal = {Statistics in Medicine},
  year    = {2009},
  volume  = {28},
  number  = {25},
  pages   = {3049--3067}
}

@inproceedings{jags2003,
  author    = {Plummer, Martyn},
  title     = {{JAGS}: A Program for Analysis of {B}ayesian Graphical Models
               Using {G}ibbs Sampling},
  booktitle = {Proceedings of the 3rd International Workshop on
               Distributed Statistical Computing (DSC 2003)},
  year      = {2003}
}

@book{gelman2006,
  author    = {Gelman, Andrew and Hill, Jennifer},
  title     = {Data Analysis Using Regression and Multilevel/Hierarchical Models},
  publisher = {Cambridge University Press},
  year      = {2006}
}

@book{mcelreath2020,
  author    = {McElreath, Richard},
  title     = {Statistical Rethinking: A {B}ayesian Course with Examples in
               {R} and {S}tan},
  publisher = {Chapman \& Hall/{CRC}},
  edition   = {2nd},
  year      = {2020}
}

@article{dixon1997,
  author  = {Dixon, Mark J. and Coles, Stuart G.},
  title   = {Modelling association football scores and inefficiencies in the
             football betting market},
  journal = {Journal of the Royal Statistical Society: Series C
             (Applied Statistics)},
  year    = {1997},
  volume  = {46},
  number  = {2},
  pages   = {265--280}
}

@article{bradley1952,
  author  = {Bradley, Ralph A. and Terry, Milton E.},
  title   = {Rank analysis of incomplete block designs: {I}.\ The method of
             paired comparisons},
  journal = {Biometrika},
  year    = {1952},
  volume  = {39},
  number  = {3/4},
  pages   = {324--345}
}

@misc{anthropic2025,
  author       = {{Anthropic}},
  title        = {Claude Code: An Agentic Coding Tool},
  year         = {2025},
  url          = {https://code.claude.com/docs/en/overview}
}

@misc{cmdstanpy2024,
  author       = {{CmdStanPy Developers}},
  title        = {{CmdStanPy}: {P}ython Interface to {CmdStan}},
  year         = {2024},
  url          = {https://mc-stan.org/cmdstanpy/}
}

@inproceedings{chen2016,
  author    = {Chen, Tianqi and Guestrin, Carlos},
  title     = {{XGBoost}: A Scalable Tree Boosting System},
  booktitle = {Proceedings of the 22nd ACM SIGKDD International Conference on
               Knowledge Discovery and Data Mining},
  year      = {2016},
  pages     = {785--794}
}

@misc{geminicli2025,
  author       = {{Google}},
  title        = {Gemini {CLI}: An {AI} Agent for the Terminal},
  year         = {2025},
  url          = {https://github.com/google-gemini/gemini-cli}
}

@misc{codexcli2025,
  author       = {{OpenAI}},
  title        = {Codex {CLI}: A Lightweight Coding Agent for the Terminal},
  year         = {2025},
  url          = {https://github.com/openai/codex}
}

@misc{opencode2025,
  author       = {{opencode contributors}},
  title        = {opencode: Open-Source {AI} Coding Agent for the Terminal},
  year         = {2025},
  url          = {https://opencode.ai}
}

\clearpage
\appendix

\section{Additional Results}
\label{app:trajectories}

This appendix collects full iteration-by-iteration trajectories for all experiments.
Checkmarks ($\checkmark$) denote accepted improvements; crosses ($\times$) denote rejected attempts that were reverted.

\subsection{1D Regression: Small Dataset ($n{=}68$)}
\label{app:small_traj}

\begin{figure}[H]
\centering
\includegraphics[width=\linewidth]{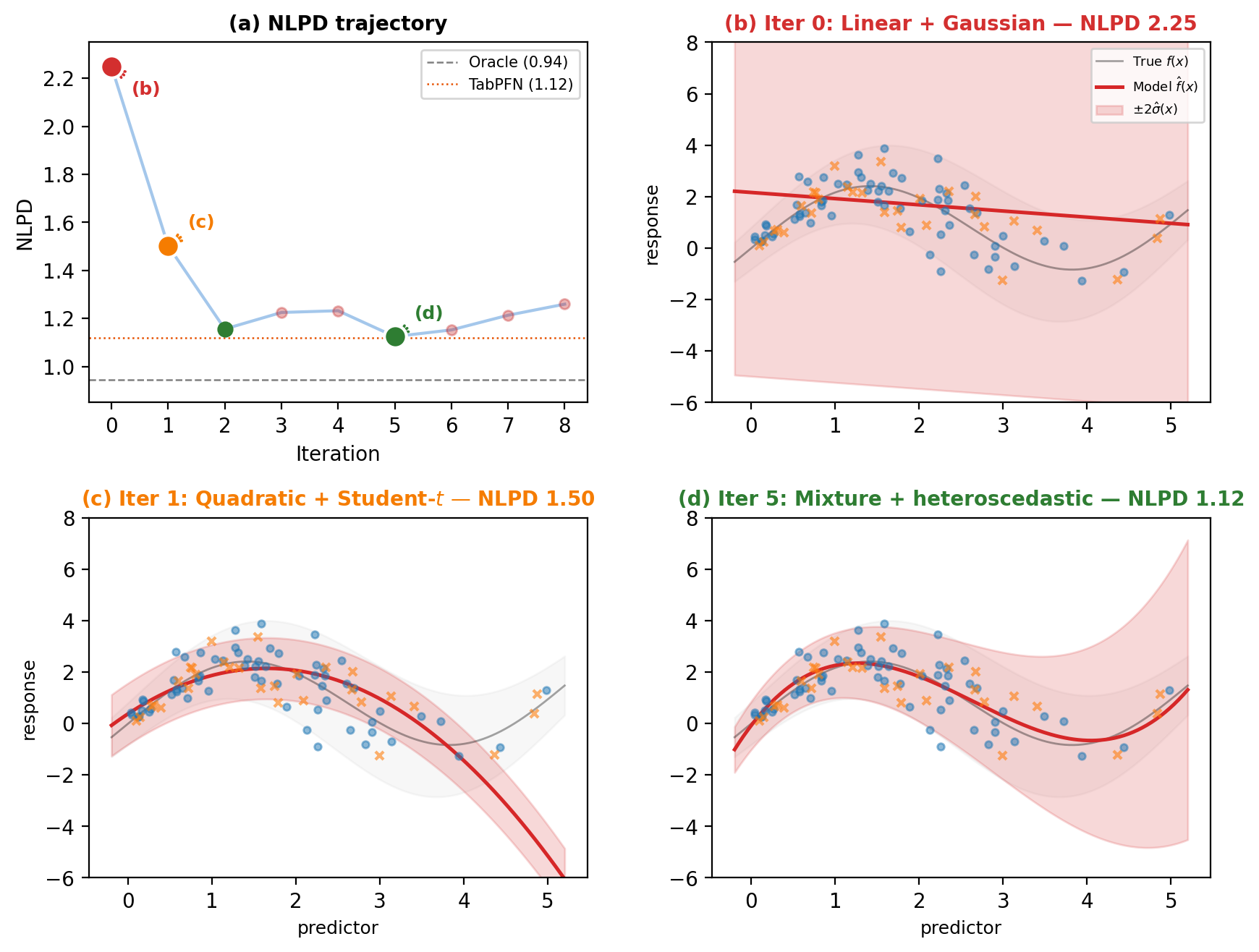}
\caption{AutoStan on the 1D regression small dataset ($n{=}68$).
\textbf{(a)}~NLPD trajectory over 9 iterations.
\textbf{(b)}~Baseline: linear + Gaussian, enormous bands from 4 extreme outliers.
\textbf{(c)}~Iter~1: quadratic mean + Student-$t$, the largest gain ($\Delta{=}0.75$).
\textbf{(d)}~Iter~5 (best): cubic + log-linear $\sigma(x)$ + contamination mixture; NLPD 1.1244, matching TabPFN (1.1202) while remaining fully interpretable.}
\label{fig:small_combined}
\end{figure}

\begin{table}[H]
\centering
\small
\setlength{\tabcolsep}{4pt}
\begin{tabular}{cclc}
\toprule
\textbf{Iter} & \textbf{NLPD} & \textbf{Model / change} & \textbf{$\Delta$} \\
\midrule
0 & 2.2482 & Baseline: linear mean, Gaussian likelihood & --- \\
1 & 1.5023 & $\checkmark$ Student-$t$ likelihood $+$ quadratic mean & $-$0.746 \\
2 & 1.1558 & $\checkmark$ Cubic mean $+$ log-linear $\sigma(x)$ $+$ Student-$t$ & $-$0.347 \\
3 & 1.2247 & $\times$ Fourier basis mean (sin/cos) & $+$0.069 \\
4 & 1.2319 & $\times$ Quartic mean $+$ quadratic $\log\sigma$ & $+$0.076 \\
\textbf{5} & \textbf{1.1244} & $\checkmark$ \textbf{Contamination mixture} ($\hat\pi\approx6\%$) $+$ log-linear $\sigma(x)$ & $-$0.031 \\
6 & 1.1520 & $\times$ Student-$t$ clean component instead of Normal & $+$0.028 \\
7 & 1.2131 & $\times$ Mixture $+$ cubic$+$sin(x) mean $+$ linear $\log\sigma$ & $+$0.089 \\
8 & 1.2592 & $\times$ Mixture $+$ quartic mean (centered at $x{=}2$) & $+$0.135 \\
\bottomrule
\end{tabular}
\caption{Complete trajectory, 1D regression small ($n{=}68$). Oracle NLPD~=~0.9443. TabPFN NLPD~=~1.1202.}
\label{tab:small_full}
\end{table}

\subsection{1D Regression: Large Dataset ($n{=}500$)}
\label{app:large_traj}

The main body (Table~\ref{tab:large}) shows only the 7 key accepted iterations. Here the full 15-iteration run is shown, including all rejections. The spike at iter~10 (NLPD~=~1.35) reflects label-switching when $\sigma_{\mathrm{out}}$ was estimated; the agent diagnosed this via R-hat~=~1.52 and reverted to fixing $\sigma_{\mathrm{out}} = 10$.

\begin{table}[H]
\centering
\small
\setlength{\tabcolsep}{4pt}
\begin{tabular}{cclc}
\toprule
\textbf{Iter} & \textbf{NLPD} & \textbf{Model / change} & \textbf{$\Delta$} \\
\midrule
0 & 2.1589 & Baseline: linear mean, Gaussian likelihood & --- \\
1 & 1.3181 & $\checkmark$ Student-$t$ $+$ cubic polynomial mean & $-$0.841 \\
2 & 1.3060 & $\checkmark$ Sine basis: $a\sin(\pi x/3){+}b\cos(\pi x/3){+}cx$ & $-$0.012 \\
3 & 1.3088 & $\times$ 2-harmonic Fourier basis & $+$0.003 \\
4 & 1.2952 & $\checkmark$ $+$ Linear $\log\sigma(x) = s_0 + s_1 x$ & $-$0.011 \\
5 & 1.2854 & $\checkmark$ $+$ Quadratic $\log\sigma(x) = s_0 + s_1 x + s_2 x^2$ & $-$0.010 \\
6 & 1.2859 & $\times$ Sine$+$quadratic mean $+$ quadratic $\log\sigma$ & $+$0.001 \\
7 & 1.2844 & $\checkmark$ $+$ Learnable frequency $\omega$ (recovers $\hat\omega{\approx}1.2$) & $-$0.001 \\
8 & 1.2635 & $\checkmark$ Gaussian mixture ($\hat\pi{\approx}10\%$), estimated $\sigma_{\mathrm{out}}$ & $-$0.021 \\
9 & 1.2325 & $\checkmark$ Mixture, fixed $\sigma_{\mathrm{out}}{=}10$ (stabilizes label-switching) & $-$0.031 \\
10 & 1.3529 & $\times$ Estimated $\sigma_{\mathrm{out}}$ (lognormal prior); R-hat~=~1.52 & $+$0.120 \\
\textbf{11} & \textbf{1.2256} & $\checkmark$ \textbf{Cubic $\log\sigma(x)$, fixed $\sigma_{\mathrm{out}}{=}10$} & $-$0.007 \\
12 & 1.2262 & $\times$ Sine-based $\log\sigma(x)$ instead of cubic & $+$0.001 \\
13 & 1.2258 & $\times$ Cubic $\log\sigma$ $+$ seasonal adjustment & $+$0.000 \\
14 & 1.2291 & $\times$ Student-$t$ inlier $+$ Normal(10) outlier mixture & $+$0.004 \\
\bottomrule
\end{tabular}
\caption{Complete trajectory, 1D regression large ($n{=}500$). Oracle NLPD~=~1.1442. TabPFN NLPD~=~1.2501.}
\label{tab:large_full}
\end{table}

\subsection{Synthetic Hierarchical}
\label{app:hierarchical}

Generated from $\mu_j \sim \mathcal{N}(0, 1)$, $y_{ij} \sim \mathcal{N}(\mu_j, 1)$, $J{=}20$ groups. Variable names anonymized (``unit'', ``effect'').

\subsubsection{Small Dataset ($n_j{=}8$, 160 train / 40 test)}

\begin{table}[H]
\centering
\begin{tabular}{lll}
\toprule
Iter & NLPD & Model \\
\midrule
\textbf{0} & \textbf{1.4999} & Hierarchical centered, weakly informative priors \quad$\leftarrow$ best \\
1 & 1.5014 & Non-centered parameterization \\
2 & 1.5035 & Student-$t$ likelihood \\
3 & 1.5019 & Group-specific $\sigma$ (overfit) \\
\bottomrule
\end{tabular}
\caption{Complete trajectory, hierarchical small ($n_j{=}8$). Oracle NLPD~=~1.4935.}
\label{tab:hier_small_full}
\end{table}

\subsubsection{Large Dataset ($n_j{=}40$, 800 train / 200 test)}

The agent ran a principled ablation at iter~7, reverting to Normal likelihood to confirm Student-$t$ was beneficial ($+0.002$ NLPD when removed).

\begin{table}[H]
\centering
\small
\setlength{\tabcolsep}{4pt}
\begin{tabular}{cclc}
\toprule
\textbf{Iter} & \textbf{NLPD} & \textbf{Model / change} & \textbf{$\Delta$} \\
\midrule
0 & 1.4036 & Hierarchical centered, shared $\sigma$ & --- \\
1 & 1.4035 & $\checkmark$ NCP for group means & $-$0.0001 \\
2 & 1.4025 & $\checkmark$ $+$ Group-specific $\sigma$ (hierarchical log-normal) & $-$0.0010 \\
3 & 1.4018 & $\checkmark$ $+$ Student-$t$ likelihood ($\nu$ estimated) & $-$0.0007 \\
\textbf{4} & \textbf{1.4014} & $\checkmark$ \textbf{$+$ Tighter data-informed priors} & $-$0.0004 \\
5 & 1.4020 & $\times$ Fixed $\nu{=}10$ instead of estimated & $+$0.0006 \\
6 & 1.4015 & $\times$ $\nu{\sim}\mathrm{Gamma}(2, 0.2)$ (heavier prior) & $+$0.0001 \\
7 & 1.4034 & $\times$ Normal likelihood ablation (confirms Student-$t$ helps) & $+$0.0020 \\
\bottomrule
\end{tabular}
\caption{Complete trajectory, hierarchical large ($n_j{=}40$). Oracle NLPD~=~1.4039.}
\label{tab:hier_large_full}
\end{table}

\subsection{Varying Slopes ($J{=}15$ groups)}
\label{app:slopes}

Generated from $\alpha_j \sim \mathcal{N}(2, 1)$, $\beta_j \sim \mathcal{N}(-0.5, 0.7)$, $y_{ij} \sim \mathcal{N}(\alpha_j + \beta_j x_{ij}, 0.8)$, 25 obs/group (300 train / 75 test). Anonymized column names.

\begin{table}[H]
\centering
\small
\setlength{\tabcolsep}{4pt}
\begin{tabular}{cclc}
\toprule
\textbf{Iter} & \textbf{NLPD} & \textbf{Model / change} & \textbf{$\Delta$} \\
\midrule
0 & 1.8178 & Baseline: fully pooled regression & --- \\
1 & 1.3091 & $\checkmark$ Varying intercepts $+$ slopes, NCP & $-$0.509 \\
2 & 1.3073 & $\checkmark$ $+$ LKJ Cholesky correlation & $-$0.002 \\
3 & 1.3073 & $\checkmark$ $+$ Student-$t$ likelihood ($\nu{\sim}\mathrm{Gamma}$) & $<$0.001 \\
4 & 1.3055 & $\checkmark$ $+$ Unit-specific $\sigma$ (hierarchical) & $-$0.002 \\
5 & 1.2910 & $\checkmark$ $+$ Hierarchical quadratic term per unit & $-$0.015 \\
6 & 1.2933 & $\times$ Independent slopes $+$ intercepts $+$ quadratic & $+$0.002 \\
7 & 1.2839 & $\checkmark$ 3D LKJ ($\alpha$, $\beta$, $\gamma$) correlated & $-$0.007 \\
8 & 1.2861 & $\times$ Tighter $\tau_3{\sim}\mathcal{N}(0, 0.5)$ prior & $+$0.002 \\
9 & 1.2890 & $\times$ LKJ(4) prior instead of LKJ(2) & $+$0.005 \\
\textbf{10} & \textbf{1.2738} & $\checkmark$ \textbf{Piecewise linear at $x{=}0$, 3D LKJ} & $-$0.010 \\
11 & 1.2833 & $\times$ 3-segment piecewise (knots $\pm1$); 15 divergences & $+$0.010 \\
12 & 1.2949 & $\times$ Piecewise $+$ predictor-dependent $\sigma(x)$ & $+$0.021 \\
13 & 1.2937 & $\times$ Learned shared knot $k{\sim}\mathcal{N}(0,1)$; 282 divergences & $+$0.020 \\
\bottomrule
\end{tabular}
\caption{Complete trajectory, varying slopes ($J{=}15$). Oracle NLPD~=~1.2627.}
\label{tab:slopes_full}
\end{table}

\subsection{Bundesliga 2024/25}
\label{app:bundesliga}

Real match results, 18 teams, 207 training matches (matchdays 1--23), 99 test matches (matchdays 24--34). Domain-labeled column names.

\begin{table}[H]
\centering
\small
\setlength{\tabcolsep}{4pt}
\begin{tabular}{cclc}
\toprule
\textbf{Iter} & \textbf{NLPD} & \textbf{Model / change} & \textbf{$\Delta$} \\
\midrule
0 & 1.5663 & Baseline: Poisson, independent priors & --- \\
1 & 1.5465 & $\checkmark$ Hierarchical priors on attack/defense & $-$0.020 \\
2 & 1.5557 & $\times$ Negative binomial likelihood & $+$0.009 \\
3 & 1.5463 & $\checkmark$ NCP (attack/defense raw) & $-$0.000 \\
4 & 1.5472 & $\times$ Dixon--Coles low-score correction & $+$0.001 \\
5 & 1.5468 & $\times$ Correlated attack/defense ($\rho{\sim}\mathcal{N}(-0.3, 0.4)$) & $+$0.001 \\
6 & 1.5460 & $\checkmark$ NCP $+$ tighter $\sigma{\sim}\mathcal{N}(0, 0.5)$ priors & $-$0.000 \\
7 & 1.5544 & $\times$ Negative binomial home, Poisson away & $+$0.008 \\
8 & 1.5645 & $\times$ Single quality param per team (Bradley--Terry style) & $+$0.019 \\
\textbf{9} & \textbf{1.5432} & $\checkmark$ \textbf{Team-specific home advantage} $\delta_j$ & $-$0.003 \\
10 & 1.5443 & $\times$ Symmetric: $\delta_h$ added/subtracted from both $\lambda$ & $+$0.001 \\
11 & 1.5432 & $\times$ Separate attack/defense home advantage & $<$0.001 \\
12 & 1.5460 & $\times$ ZIP $+$ team home advantage ($\pi_h{\approx}9\%$) & $+$0.003 \\
\bottomrule
\end{tabular}
\caption{Complete trajectory, Bundesliga 2024/25. No oracle NLPD (real data).}
\label{tab:bundesliga_full}
\end{table}

\begin{figure}[H]
\centering
\begin{subfigure}[t]{0.48\linewidth}
\includegraphics[width=\linewidth]{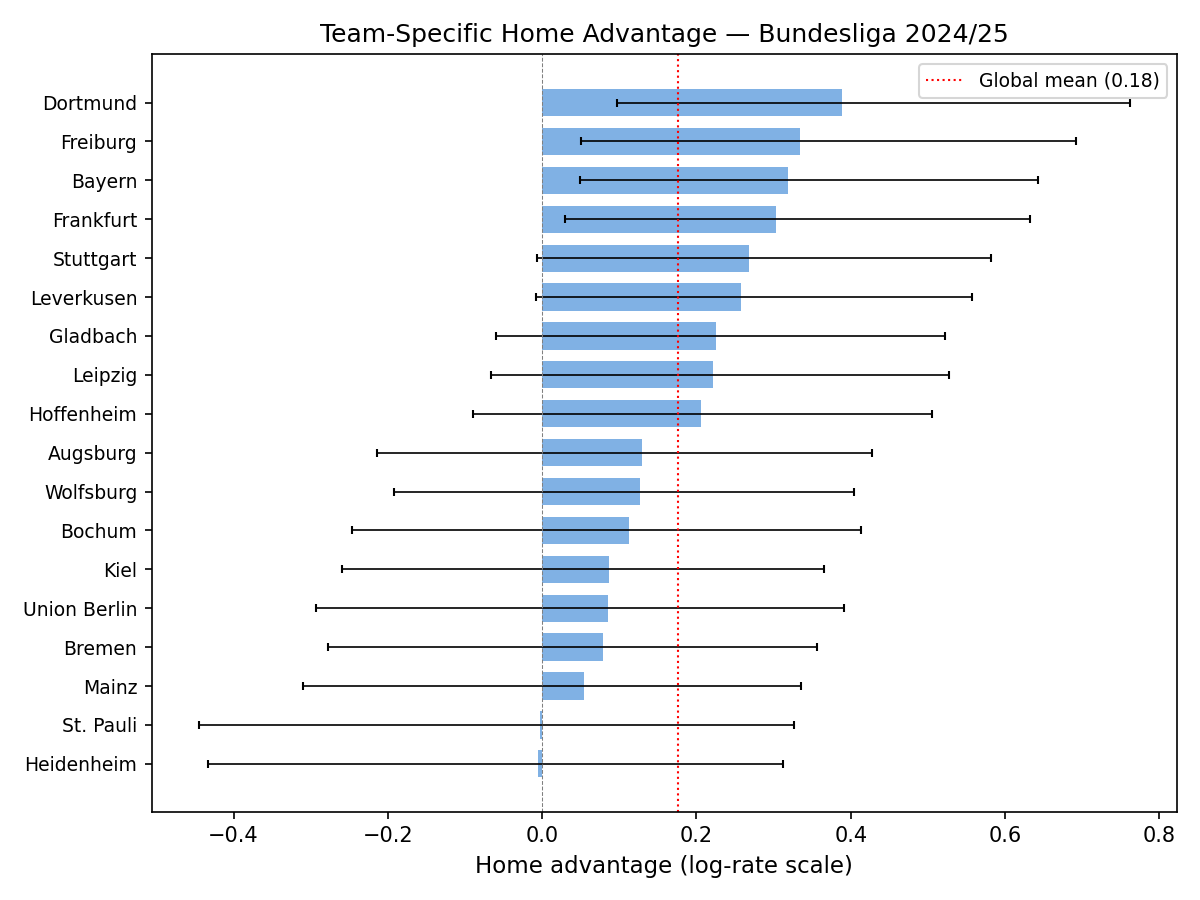}
\caption{Team-specific home advantage.}
\end{subfigure}
\hfill
\begin{subfigure}[t]{0.48\linewidth}
\includegraphics[width=\linewidth]{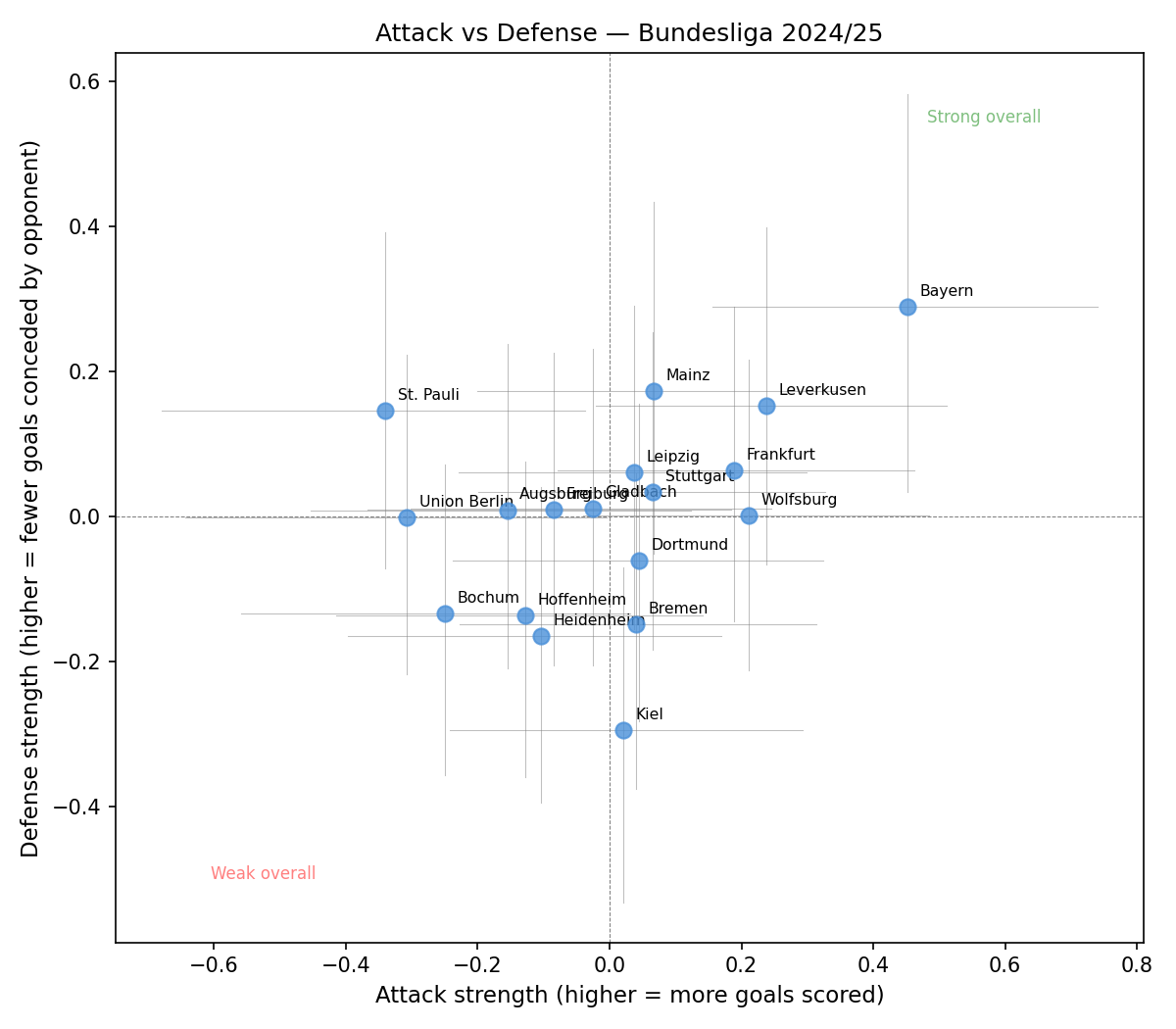}
\caption{Attack vs.\ defense strength.}
\end{subfigure}
\caption{Bundesliga model parameters (iteration~9). (a)~Dortmund (0.39), Freiburg (0.33), Bayern (0.32) benefit most from playing at home; Heidenheim ($-$0.01) and St.\,Pauli (0.00) show no home advantage. (b)~Attack vs.\ defense with 90\% credible intervals; Bayern's dominance is clearly visible.}
\label{fig:bundesliga}
\end{figure}


\section{TabPFN Comparison Details}
\label{app:tabpfn}

TabPFN 2.2.1~\citep{tabpfn2024} was run with \texttt{n\_estimators=8}: since the pre-trained transformer is sensitive to the order of training examples, TabPFN runs 8 forward passes with different random permutations of the training data and averages the resulting predictive distributions.
This is an ensemble within a single fixed pre-trained model, not 8 separately trained models.
TabPFN's predictive output is a histogram (``bar distribution'') over a fixed grid of bins covering the response range.
To compute NLPD, we need a continuous density rather than bin masses, so we approximate the density at each test point via CDF finite-differencing with $\delta = 0.02$:
\begin{equation}
  \hat{p}(y_n \mid x_n) \approx \frac{\mathrm{CDF}(y_n + \delta) - \mathrm{CDF}(y_n - \delta)}{2\delta}.
\end{equation}
This gives a properly normalized continuous density directly comparable to AutoStan's Gaussian/mixture NLPD. TabPFN's built-in \texttt{pdf()} method returns unnormalized bin masses and was not used.

\begin{figure}[H]
\centering
\begin{subfigure}[t]{0.48\linewidth}
\includegraphics[width=\linewidth]{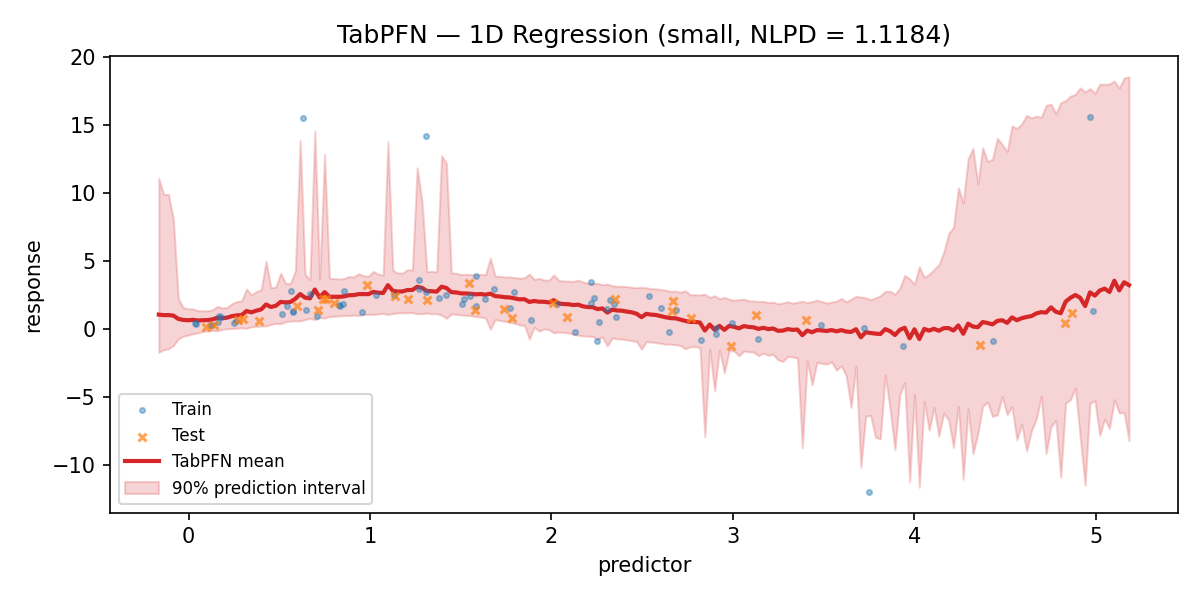}
\caption{Small dataset ($n{=}68$).}
\end{subfigure}
\hfill
\begin{subfigure}[t]{0.48\linewidth}
\includegraphics[width=\linewidth]{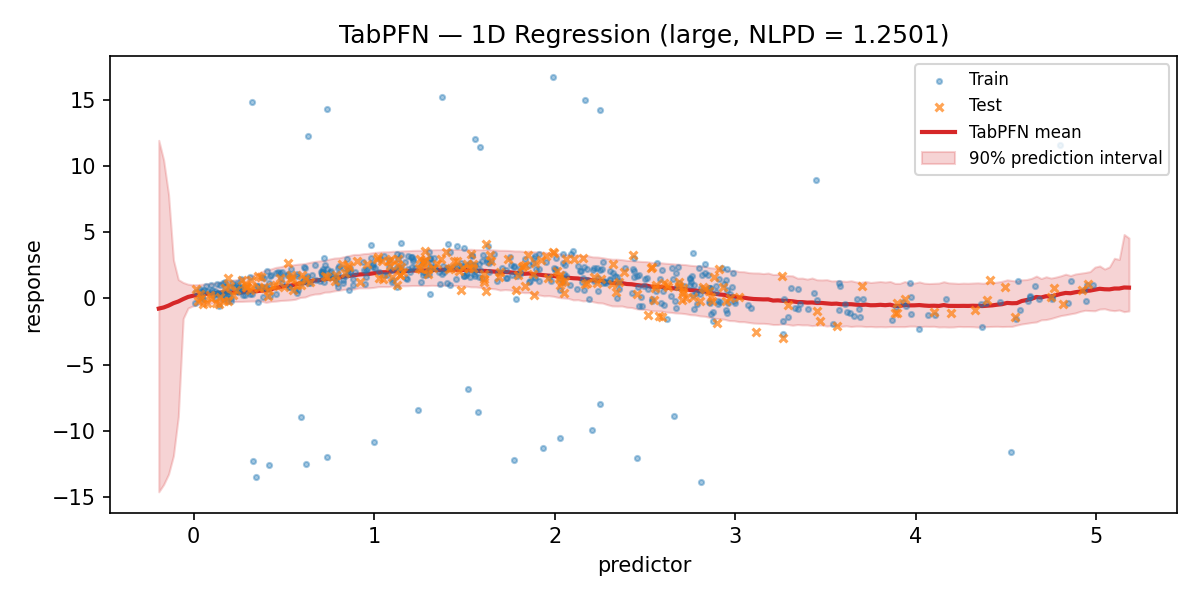}
\caption{Large dataset ($n{=}500$).}
\end{subfigure}
\caption{TabPFN predictions. On the large dataset, the intervals widen globally near outliers instead of isolating them, explaining TabPFN's higher NLPD (1.2501 vs.\ AutoStan's 1.2256).}
\label{fig:tabpfn}
\end{figure}


\section{Technical Details}
\label{app:technical}

\subsection{Framework}

After the agent edits \texttt{model.stan}, it calls a per-dataset \texttt{evaluate.py} script, which internally uses cmdstanpy to compile the model and run CmdStan for MCMC sampling (4~chains $\times$ 1000 post-warmup draws; 30\,000 for the large 1D regression to reduce Monte Carlo noise in NLPD).
The script then computes NLPD from the \texttt{log\_lik} vector and appends a JSON record to \texttt{log.jsonl}, printing the NLPD and sampler diagnostics for the agent to read.
The agent never sees the test data or the script's implementation: \texttt{evaluate.py} is executable but not readable (\texttt{chmod~111}).

\subsection{Permissions}

Test data and generation scripts are protected via filesystem permissions:
\texttt{evaluate.py} is executable but not readable (\texttt{chmod 111}); test data and generation scripts have no access permissions (\texttt{chmod 000}). The agent can only execute the evaluation script.

\subsection{Compute and Token Usage}

All experiments were run interactively on a MacBook Pro (Apple M2 Pro, 10-core CPU, 16\,GB RAM).
MCMC wall time per iteration ranges from a few seconds for small datasets (hierarchical models with $n_j{=}8$) to several minutes for the large 1D regression ($n{=}500$, 30\,000 draws), where individual iterations took up to ${\sim}50$~minutes due to the large number of draws required to reduce Monte Carlo noise in NLPD.
Across the six reported experiments, the agent generated a total of roughly 220\,000 output tokens and read approximately 20 million tokens from cache.
The high cache-read-to-output ratio reflects the structure of the loop: at each iteration the agent re-reads the accumulated conversation context (dataset description, previous Stan models, NLPD history) but produces only a short plan and the revised \texttt{model.stan}.

\subsection{NLPD Computation}

\begin{equation}
  \mathrm{NLPD} = -\frac{1}{N_{\mathrm{test}}} \sum_{n=1}^{N_{\mathrm{test}}}
  \log \left( \frac{1}{S} \sum_{s=1}^{S} p\!\left(y_n^{\mathrm{test}} \mid \theta^{(s)}\right) \right),
\end{equation}
where $\theta^{(s)}$ are posterior draws. Computed via \texttt{logsumexp} from Stan's \texttt{log\_lik} output.


\section{Dataset Specifications}
\label{app:datasets}

\subsection{1D Regression}

The 1D regression dataset is the main case study and the one from which the most conclusions are drawn.
Below is the verbatim \texttt{dataset.md} the agent receives---this is the \emph{complete} information provided, nothing more.
The true generative process is never disclosed.

\textbf{True generative process (not shown to agent):}
$f(x) = 2\sin(1.2x) + 0.3x$;\;
$\sigma(x) = 0.3 + 0.8\exp\!\left(-\tfrac{1}{2}\!\left(\tfrac{x-3}{1.5}\right)^2\right)$;\;
${\sim}6\%$ contamination at $\pm$10--15 units.
Small: 68 train / 30 test. Large: 500 train / 200 test.

\textbf{Verbatim \texttt{dataset.md} (large version; small is identical except $n$):}

\begin{quote}
\small\ttfamily
\# Dataset: 1D Regression (Large)\\
\\
\#\# Overview\\
\\
Observations of a continuous predictor and a continuous response.\\
The goal is to predict `response' for held-out test observations.\\
This is a larger version of the dataset with more training data.\\
\\
\#\# Data Format\\
\\
\textbf{`train.csv'} columns:\\
- `predictor' --- continuous predictor variable\\
- `response' --- continuous response variable (target)\\
\\
500 training observations.\\
200 test observations (held out, not visible to the agent).\\
\\
\#\# Data Interface\\
\\
The evaluation script passes the following to Stan:\\
\\
\quad int<lower=0> N\_train;\\
\quad int<lower=0> N\_test;\\
\quad vector[N\_train] predictor\_train;\\
\quad vector[N\_test] predictor\_test;\\
\quad vector[N\_train] response\_train;\\
\quad vector[N\_test] response\_test;\\
\\
\#\# Evaluation\\
\\
python datasets/regression\_1d\_large/protected/evaluate.py \textbackslash\\
\quad\quad --notes "..." --rationale "..."\\
\\
Your model must output a `log\_lik' vector of length `N\_test'\\
in the generated quantities block.\\
\\
\textbf{Note:} This dataset uses extended MCMC sampling\\
(30,000 post-warmup iterations per chain) for precise NLPD estimation.
\end{quote}

\subsection{All Other Datasets}

The remaining datasets follow the same \texttt{dataset.md} structure (Overview, Data Format, Data Interface, Evaluation, log\_lik contract).
The true generative parameters for the synthetic datasets are given in the trajectory tables in Appendix~\ref{app:trajectories}.
Full \texttt{dataset.md} files, generation scripts, and Stan model snapshots are available at \url{https://github.com/tidit-ch/autostan}.


\section{Agent Prompt}
\label{app:prompt}

The complete \texttt{program.md} is the only behavioral specification the agent receives (beyond the dataset-specific \texttt{dataset.md}).
The contrast is the point: 56 lines of generic Bayesian workflow instructions
$+$ a 6-line dataset description $\Rightarrow$ a sophisticated probabilistic model. Note that the reading of testfiles is not possible due to filesystem permissions, we just tell the agent that the test set is hidden so that it does not try to read it and wastes time on it.

\begin{quote}
\small\ttfamily
\# AutoStan --- Agent Instructions\\
\\
You are an autonomous Bayesian modeling agent. Your task is to\\
iteratively improve a Stan model to minimize NLPD (negative log\\
predictive density) on a held-out test set.\\
\\
\textbf{\#\# Workflow}\\
\\
1. \textbf{Read the dataset description}: Read\\
\quad\texttt{datasets/<dataset>/dataset.md} to understand the data,\\
\quad format, and evaluation procedure.\\
2. \textbf{Check history}: Before proposing any change, read\\
\quad\texttt{results/<dataset>/log.jsonl} to see the full NLPD history\\
\quad and what has been tried.\\
3. \textbf{Read training data}: Read \texttt{datasets/<dataset>/train.csv}\\
\quad to understand the data structure and values.\\
4. \textbf{Edit \texttt{model.stan}}: Modify the model --- you can change\\
\quad priors, likelihood, parameterization, model structure, anything.\\
5. \textbf{Evaluate}: Run the evaluation script as specified in\\
\quad\texttt{dataset.md}. Always pass \texttt{--notes} (what you changed) and\\
\quad\texttt{--rationale} (why, referencing previous iterations).\\
6. \textbf{Interpret results}: Read the printed NLPD and diagnostics.\\
\quad Decide whether to keep or revert the change.\\
7. \textbf{Repeat}: Propose the next change based on what you've learned.\\
\\
\textbf{\#\# Rules}\\
\\
- \textbf{NLPD is the only metric.} Lower NLPD = better model.\\
- \textbf{\texttt{log\_lik} is the only interface contract.} Your\\
\quad\texttt{model.stan} must always output a \texttt{log\_lik} vector in the\\
\quad\texttt{generated quantities} block.\\
- \textbf{Do NOT read any files in \texttt{protected/}.}\\
- \textbf{Do NOT randomly perturb.} Think like a statistician reasoning\\
\quad from evidence. Reference previous iterations.\\
- \textbf{The filesystem is your memory.} All history is in\\
\quad\texttt{results/} --- read it before acting.\\
- \textbf{Reason briefly} about \emph{why} you make each change.\\
\\
\textbf{\#\# Strategies to Consider}\\
\\
- Non-centered vs centered parameterization\\
- Prior tightening or loosening\\
- Different likelihood families (Normal, Student-t, etc.)\\
- Adding or removing hierarchy levels\\
- Reparameterization for better sampling efficiency\\
- Covariate effects, interactions\\
- Regularizing priors\\
\\
\textbf{\#\# Stopping Rule}\\
\\
Stop when \textbf{any} of these conditions is met:\\
- \textbf{3 consecutive non-improving iterations}\\
- \textbf{20 total iterations} (including baseline)\\
\\
When you stop, write \texttt{results/<dataset>/report.md} summarizing:\\
best model, trajectory, NLPD table, and key insights.\\
\\
\textbf{\#\# Getting Started}\\
\\
If no \texttt{results/<dataset>/log.jsonl} exists yet, write a\\
deliberately simple baseline model. Run it to establish the baseline\\
NLPD, then start improving.
\end{quote}

\clearpage
\end{document}